\documentclass[letterpaper, 10 pt, conference]{ieeeconf}  

\IEEEoverridecommandlockouts                              

\usepackage{listings}
\usepackage{amsthm}

\usepackage{amsmath}
\usepackage{amssymb}
\usepackage{breqn}
\usepackage{mathtools,algorithm}
\usepackage[noend]{algpseudocode}
\usepackage{textcomp}
\usepackage{gensymb}
\usepackage{graphicx}
\usepackage{color}
\usepackage{varwidth}
\usepackage{microtype}
\usepackage{acronym}
\usepackage{hyperref}
\usepackage{multirow}
\usepackage{cite}

\usepackage[noend]{algpseudocode}
\usepackage[symbol]{footmisc}
\newtheorem{theorem}{Theorem}

\newtheorem{lemma}[theorem]{Lemma}

\usepackage[mathscr]{euscript}
\title{\LARGE \bf
Model Predictive Actor-Critic: Accelerating Robot \\ Skill Acquisition with Deep Reinforcement Learning 
}

\author
{Andrew S. Morgan$^{1}$*, Daljeet Nandha$^{2}$*, Georgia Chalvatzaki$^{2}$, \\ Carlo D'Eramo$^{2}$, Aaron M. Dollar$^{1}$, and Jan Peters$^{2}$
\thanks{Computations performed on the Lichtenberg cluster of TU Darmstadt and the Yale HPC Grace cluster. This work has been partially funded by the RoboTrust, Skill4Robots projects and NSF Grants IIS-1752134 \& IIS-1900681. Dr. Chalvatzaki is funded by the DFG EN Program (CH 2676/1-1).}
\thanks{$^{1}$Department of Mechanical Engineering \& Materials Science, Yale University, USA. (\tt \{andrew.morgan, aaron.dollar\}@yale.edu).}
\thanks{$^2$Intelligent Autonomous Systems, Technische Universit\"{a}t Darmstadt, Germany (\tt daljeet.nandha@stud.tu-darmstadt.de, \tt  \{georgia, carlo\}@robot-learning.de, mail@jan-peters.net).}
\thanks{$^*$Authors contributed equally.}
}

\begin{document}

\acrodef{MoPAC}{\emph{Model Predictive Actor-Critic}}
\acrodef{RL}{\emph{Reinforcement Learning}}
\acrodef{PPO}{\emph{Proximal Policy Optimization}}
\acrodef{SAC}{\emph{Soft-Actor Critic}}
\acrodef{DDPG}{\emph{Deep Deterministic Policy Gradient}}
\acrodef{MDP}{\emph{Markov Decision Process}}
\acrodef{MPC}{\emph{Model Predictive Control}}
\acrodef{MPPI}{\emph{Model Predictive Path Integral}}
\acrodef{MAAC}{\emph{Model-Augmented Actor-Critic}}
\acrodef{MAML}{\emph{Model-Agnostic Meta-Learning}}
\acrodef{MB-MPO}{\emph{Model-Based Meta-Policy-Optimization}}
\acrodef{MBPO}{\emph{Model-Based Policy-Optimization}}
\acrodef{MLP}{\emph{Multi-Layer Perceptron}}
\acrodef{TRPO}{\emph{Trust-Region Policy Optimization}}
\acrodef{ME-TRPO}{\emph{Model-Ensemble Trust-Region Policy Optimization}}
\acrodef{PETS}{\emph{Probabilistic Ensembles with Trajectory Sampling}}
\acrodef{DQL}{\emph{Deep Q-Learning}}
\acrodef{DMPC}{\emph{Deep Value Model Predictive Control}}
\acrodef{POLO}{\emph{Plan Online and Learn Offline}}
\acrodef{MPQ}{\emph{Model Predictive Q-Learning}}
\acrodef{i-MPPI}{\emph{Information theoretic MPPI}}
\acrodef{MBRL}{\emph{Model-Based Reinforcement Learning}}

\maketitle
\thispagestyle{empty}
\pagestyle{empty}

\begin{abstract}

Substantial advancements to model-based reinforcement learning algorithms have been impeded by the model-bias induced by the collected data, which generally hurts performance.
Meanwhile, their inherent sample efficiency warrants utility for most robot applications, limiting potential damage to the robot and its environment during training. 
Inspired by information theoretic model predictive control and advances in deep reinforcement learning, we introduce Model Predictive Actor-Critic (MoPAC)\footnote[2]{Code available: \url{https://github.com/dnandha/mopac}.}, a hybrid model-based/model-free method that combines model predictive rollouts with policy optimization as to mitigate model bias. MoPAC leverages optimal trajectories to guide policy learning, but explores via its model-free method, allowing the algorithm to learn more expressive dynamics models. This combination guarantees optimal skill learning up to an approximation error and reduces necessary physical interaction with the environment, making it suitable for real-robot training.
We provide extensive results showcasing how our proposed method generally outperforms current state-of-the-art and conclude by evaluating MoPAC for learning on a physical robotic hand performing valve rotation and finger gaiting--a task that requires grasping, manipulation, and then regrasping of an object. 
\end{abstract}

\section{Introduction}

\begin{figure}[thpb]
      \centering
      \includegraphics[width=0.48\textwidth]{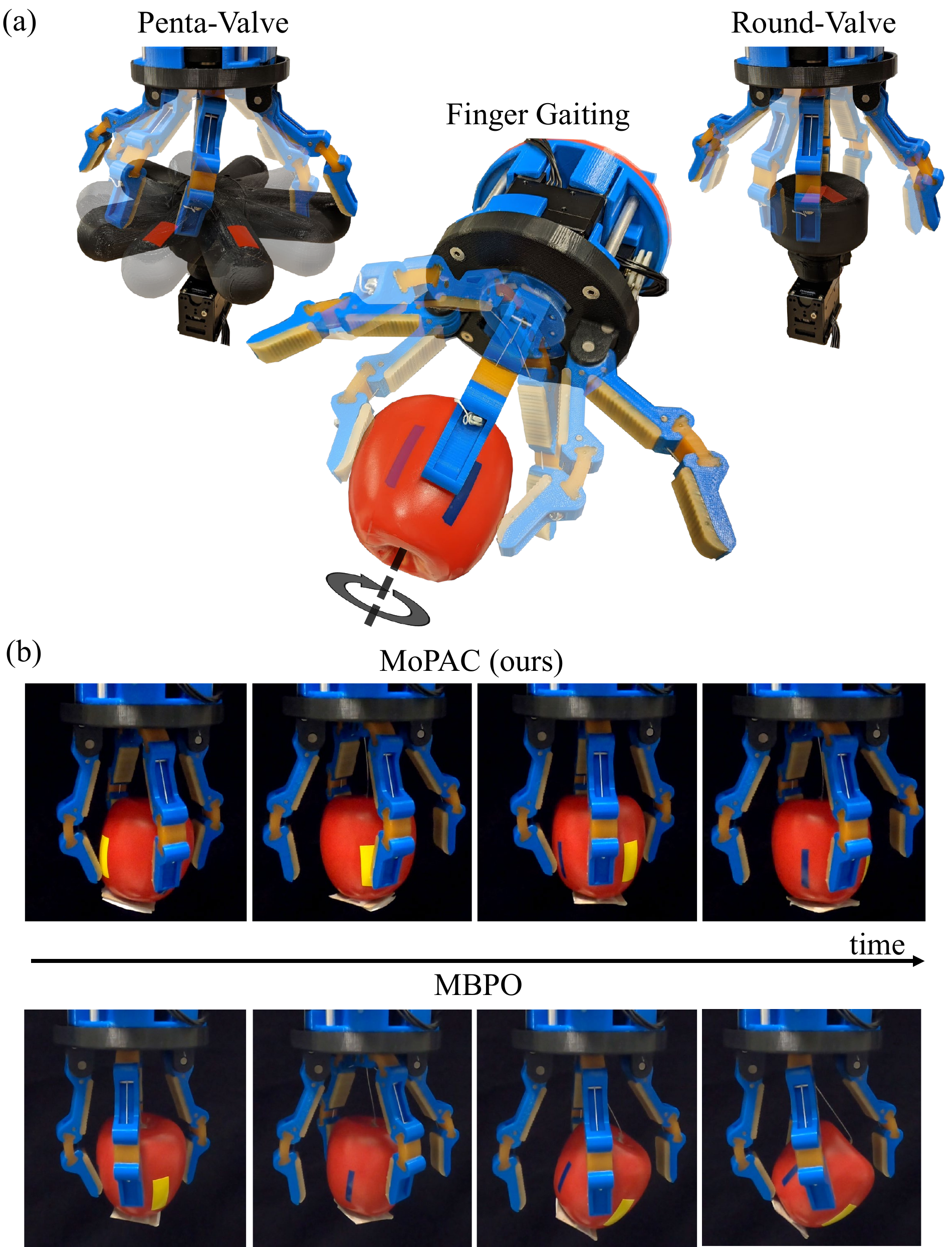}
      \vspace{-0.1cm}
      \caption{ (a) Complex robot skills, like in-hand manipulation, are difficult to acquire and perform with conventional methods. (b) MoPAC is suitable for training an underactuated hand to efficiently learn such tasks, like finger gaiting, compared to other state-of-the-art techniques.}
      \label{Splash}
      \vspace{-0.65cm}
  \end{figure}

Robotic systems are expected to operate in increasingly unstructured and dynamical environments. Aside from the difficulties evident in perception, decision making, and planning, precisely controlling robots still remains difficult, as they must operate under non-linear, contact-rich conditions. Traditional approaches to this problem include methods of optimal control \cite{tassa2008receding,atkeson1994using}, but require that a model of the robot and its environment is known \textit{a priori}, which generally cannot be guaranteed. 

In this direction, model-based reinforcement learning (MBRL) approximates iteratively the dynamics model of the environment while planning actions through trajectory optimization \cite{deisenroth2011pilco,kumar2016optimal,luo2018algorithmic,chalvatzaki2019learn,wang2019exploring, morgan2020object}, commonly using Model Predictive Control (MPC) \cite{levine2014learning, williams2017information,chua2018deep}. Though sample-efficient, MBRL has been heavily impeded by the bias in the learned model--as optimal control methods will generally continue to exploit known regions of the learned dynamics without exploring outwards to new, unrevealed states \cite{amos2018differentiable, pereira2018mpc, janner2019trust}. Model-free reinforcement learning (MFRL), on the other hand, has achieved impressive results in learning complex skills through deep neural networks \cite{williams1992simple, schulman2015trust,mnih2015human,lillicrap2015continuous,schulman2017proximal}, offering high performance. However, unlike their MB counterpart, the lack of an internal model makes MF algorithms data-hungry, suffering from poor use of samples; as the complexity of the task increases, so does the number of samples required to learn the optimal policy \cite{fujimoto2018addressing,levine2016end,haarnoja2018composable}. This disadvantage makes modern deep MFRL methods inappropriate for learning tasks on physical robots \cite{gu2017deep, haarnoja2018soft, haarnoja2018softv2}, as the increased amount of necessary interactions is likely to damage the system.



We believe that the effective combination of these RL paradigms will enable the learning of complex skills on real robots~\cite{chebotar2017combining,pong2018temporal}. In this paper, we address this challenge by investigating optimal methods of guiding policy learning, while also learning the model dynamics. Intuitively, having interweaving MB-MF components can potentially allow scheduling of separate solutions according to different phases of learning, i.e. MB when action planning is required, and MF when additional exploration is needed. 

Various works try to mix the benefits of MB and MF methods. For example, \cite{levine2013guided} proposes the use of trajectory optimization for guiding the policy search by exploring high-reward regions.
In \cite{clavera2018model}, the authors propose a MB method which learns an ensemble of models and optimizes a meta-policy objective over all models. \cite{nagabandi2018neural} proposes the initialization of MFRL algorithms from learned model dynamics, for combining the sample efficiency of MB methods with the task-specific performance of MF approaches.
Many works study the use of MB methods for accelerating the learning of MFRL. The latter can be extended into MB methods by sampling from a model (or model ensemble to increase stochasticity) \cite{kurutach2018model, chua2018deep,nagabandi2020deep}. MBPO \cite{janner2019trust} uses branched rollouts in an actor-critic setting to exploit the learned dynamics for model-based policy optimization. \cite{clavera2020model} proposes an extension to MBPO, by performing model rollouts of specific horizons, while optimizing the policy objective with back-propagation through time. 

In this work, we introduce Model Predictive Actor-Critic (MoPAC)--an algorithm that seamlessly combines the sample efficiency of MBRL, with MF actor-critic methods for improved exploration. To ensure exploitation of the learned policy, we propose the use of model predictive rollouts, a method that inherits properties of model predictive path integrals~\cite{williams2017information}, capitalizing on the free-energy of the learned system and leveraging the information theoretic constraint in the MB simulations. Our novel method is theoretically sound, as we provide a bound on the performance of trajectory optimization through MPC, when approximating the dynamics model and the value-function, that potentially allows planning for longer time-horizons. Moreover, the maximum entropy objective~\cite{haarnoja2018soft} in the policy optimization counterbalances the exploitation of model predictive rollouts with exploration on the real environment, enabling the learning of more expressive model dynamics, together with approximating the optimal policy. Our empirical results, both in simulated control tasks and on a physical robotic hand that performs in-hand manipulation, showcase the accelerated learning that MoPAC offers against representative baselines, evincing its effectiveness for learning complex skills on real-robot platforms. 

\section{Preliminaries}



We consider \textit{Reinforcement Learning} \cite{sutton2018reinforcement} for solving control problems modeled as finite-horizon Markov Decision Processes~(MDPs) $\mathcal{M} = \langle \mathcal{S}, \mathcal{A}, \mathcal{R}, \mathcal{P}, \gamma \rangle$, where $\mathcal{S}$ is the state space, $\mathcal{A}$ is the action space, $\mathcal{R}: \mathcal{S} \times \mathcal{A} \times \mathcal{S} \to \mathbb{R}$ is the reward function, $\mathcal{P}: \mathcal{S} \times \mathcal{A} \to \mathcal{S}$ is the transition kernel, and $\gamma \in [0, 1)$ is the discount factor. We define a policy $\pi \in \Pi: \mathcal{S} \times \mathcal{A} \to \mathbb{R}$ as the probability distribution of the event of executing an action $a$ in a state $s$. A policy $\pi$ induces a value function (VF) corresponding to the expected cumulative discounted reward collected by the agent when executing action $a$ in state $s$, and following the policy $\pi$ thereafter: $Q^\pi(s,a) \triangleq \mathbb{E} \left[\sum_{k=0}^\infty \gamma^k r_{i+k+1} | s_i = s, a_i = a, \pi \right]$, where $r_{i+1}$ is the reward obtained after the $i$-th transition. Solving an MDP consists of finding the optimal policy $\pi^*$, i.e. the one maximizing the expected cumulative discounted reward.

\textit{Trajectory Optimization} designs a trajectory that minimizes some measure of performance. MPC is such a technique that optimizes a cost function over a finite time-horizon, while taking into account the system dynamics. The cost function is optimized w.r.t. a control variable, yielding the optimal control value for a given state with consideration of the predicted future states. 
In essence, MPC provides a locally optimal policy or sequence of actions (up to horizon $H$), based on the following optimization problem:
\small
\begin{equation}\label{eq:mpc}
\begin{aligned}
\pi_{MPC}(s) = \arg \max_{\pi_{0:H-1}} \mathbb{E}[ \ \sum\limits_{t=0}^{H-1} \gamma^t r(s_t, a_t) + \gamma^Hr_f(s_H)]\, \\
\mbox{ } a_t=\pi_t(s_t), s_0 = s
\end{aligned}
\end{equation}
\normalsize
where the states evolve according to the transition dynamics of the MDP, i.e. $s_{t+1} = f(s_t,a_t)$. From each optimized sequence resulting from the optimization process of MPC (of length equal to horizon $H$), the first action is applied to the agent, and the procedure is repeated again at the next time step. The term $r_f(s_H)$ denotes the terminal reward. 

In the context of MBRL, \cite{farshidian2019deep} combines MB trajectory optimization with VF estimation. \cite{bhardwaj2020information} uses model predictive path integrals \cite{williams2016aggressive} in a $Q$-learning setting, and \cite{lowrey2018plan} shows that combining MPC with VF approximation yields optimal policies, however considering the true dynamics.

\section{Performance bound under imperfect model}\label{sec:bounds}
Coupling MPC with a VF, that propagates global information, optimizes action sequences for longer time horizons, while being less prominent to approximation errors than greedy action selection~\cite{lowrey2018plan}.
In this work, we design an actor-critic algorithm that will benefit from the integration of trajectory optimization for acquiring optimal control policies, together with the learning of stable VF approximations provided by modern deep actor-critic algorithms, e.g. soft actor-critic (SAC)~\cite{haarnoja2018soft}. To do so, we extend the bound of \cite{lowrey2018plan} by incorporating the approximation error when learning the dynamics model.

\begin{theorem}\label{theorem:mpc}
Let the approximation error of the dynamics model be $\epsilon_f = |\hat f(s,a) - f^*(s,a)|$, the approximation error of the VF be $\epsilon_V = max_s |\hat V(s) - V^*(s)|$, and the terminal reward of (\ref{eq:mpc}) be $r_f(s_H)=\hat V(s_H)$, then the performance of the MPC policy in (\ref{eq:mpc}) with the learned dynamics model is bounded by:
\small
\begin{equation}\label{eq:mpc_bound_2}
J(\pi^*)-J(\pi^{MPC}) \leq \frac{2\gamma^H\epsilon_{V}}{1-\gamma^H} + r_{max}\frac{1-\gamma^H}{1-\gamma} \epsilon_f.
\end{equation}
\normalsize
\end{theorem}
\noindent Proof is provided in the Appendix. As the contribution of the model error increases with the horizon, $H \to \infty$ leads to the upper bound
\small
\begin{equation}\label{eq:mpc_bound_inf}
J(\pi^*)-J(\pi^{MPC}) \leq \frac{r_{max} \epsilon_f}{1-\gamma}.
\end{equation}
\normalsize

\begin{figure}[!ht]
     \centering 
\begin{minipage}[t]{0.45\textwidth}
\begin{algorithm}[H]
\caption{\small Model Predictive Actor-Critic}\label{mopac}
\label{Alg:main}
\small
\begin{algorithmic}[1]
\State {Initialize parameters ${\theta}$, $\rho$, $\psi$, $\bar \psi$, $\phi$, $\bar \phi$}
\State {Initialize $D$, $D_{env}$, $D_{model}$} \Comment{initialize experience buffers}
\State {$\bar \phi \gets \phi$, $\bar \psi \gets \psi$} \Comment{initialize target parameters}
\For{each iteration}
\State{$D_{env} \gets D_{env} \cup \{s_{t+1},a_t,s_t\}, a_t \sim \pi_\theta(s_t)$}
\For{N epochs}
\State{Train model $f_{\rho}$ on $D_{env}$: $\rho \gets \rho - \lambda_{f}\nabla_{\rho}J_{f_{\rho}}$}
\EndFor
    \State{\textbf{end for}}
\For{M model predictive rollouts}
\State{Sample $s_t$ uniformly from $D_{env}$}
\State{Perform MPR (Alg. \ref{alg:mppi}) from $s_t$}
\State{Add transitions to $D_{model}$}
\EndFor
    \State{\textbf{end for}}
\For{G gradient steps}
\State{Update parameters using data $D\gets D_{env} \cup D_{model}$}
\State{$ \psi \gets \psi - \lambda_{\psi} \nabla_{\psi} J_{V_{\psi}}$}
\Comment{VF update}
\State{$\phi \gets \phi - \lambda_{\phi} \nabla_{\phi} J_{Q_{\phi}}$}
\Comment{Q update}
\State{$\theta \gets \theta - \lambda_{\theta} \nabla_{\theta} J_{\pi_{\theta}}$}
\Comment{policy update}
\State{$\bar \psi \gets \tau \psi + (1 - \tau)\psi$}
\Comment{target VF update}
\State{$\bar \phi \gets \tau \phi + (1 - \tau)\phi$}
\Comment{target Q update}
\EndFor
    \State{\textbf{end for}}

\EndFor
    \State{\textbf{end for}}
\end{algorithmic}
\end{algorithm}
\end{minipage}
 \vspace{-0.55cm}
\end{figure}

As expected, the performance error between the optimal policy and the MPC policy is affected by the model approximation error $\epsilon_f$, given a prediction horizon $H$. As we will show in the following, the maximum entropy exploration of SAC \cite{haarnoja2018soft} can acquire more expressive dynamics models by visiting unmodeled transitions in the environment; together with the approximation of the VF, we can leverage the bound of \eqref{eq:mpc_bound_2} to acquire near-optimal trajectories for policy learning.

\textit{Model-based monotonic improvement}, as proven by \cite{janner2019trust}, can be achieved when learning the dynamics model together with the policy. Namely, the authors give an upper-bound in the performance gain obtained when applying the learned policy to the learned dynamics model, compared to applying it on the real MDP (i.e. the true dynamics). Their finding is summarized in the following lemma.
\begin{lemma}\label{lemma:2}
Let the expected TV-distance error of the transition probability distributions be bounded by $\varepsilon_f$ and the policy divergence be bounded by $\varepsilon_{\pi}$. Then the following bound holds:
\small
\begin{equation}
\begin{aligned}
    J(\pi) - \hat J(\pi) \geq - \left[\frac{2\gamma r_{max}(\varepsilon_f+2\varepsilon_{\pi})}{1-\gamma^2} +
    \frac{4r_{max} \varepsilon_{\pi}}{1-\gamma}\right].
    \end{aligned}
\end{equation}
\end{lemma}
\normalsize
\noindent This lemma is directly applicable to our proposed algorithm. Its combination with our Theorem \ref{theorem:mpc} suggests that sufficiently low errors in model learning and policy approximation can yield near-optimal performance.

\begin{figure}
 \begin{minipage}[t]{0.45\textwidth}
\begin{algorithm}[H]
\caption{\small Model Predictive Rollouts (MPR)}\label{alg:mppi}
\small
\begin{algorithmic}[1]
\renewcommand{\algorithmicrequire}{\textbf{Input:}}
\renewcommand{\algorithmicensure}{\textbf{Output:}}
\newcommand{\algorithmicbreak}{\textbf{break}}
\newcommand{\BREAK}{\STATE \algorithmicbreak}
\Require {$s_{0}$, $\rho$, $\bar \psi$, $\pi_{\theta}$, $n$, $\zeta$, $\gamma$, $\lambda$ }
\Comment{start state, model parameters, target VF parameters, current policy, actuation noise, annealing parameter, discount, control hyperparameter}
\Ensure {$a_{MPR}, s_{MPR}$}
\Comment{MPR trajectory}


    \State {$H \gets anneal(\zeta)$}
    \Comment{horizon length}
    \State{$\mathcal{R} \gets zeros(\cdot)$} 
    \Comment{trajectory reward}

    \For {$t= 0, ..., H-1$}
        \State{Sample an initial action from policy $a_t \sim \pi_{\theta}(s_t)$}
        \State{Sample exploration noise $\{n_0, \dots, n_{H-1}\}\sim n$}
        \State {$\mathcal{R} \gets \mathcal{R} + \gamma^t r(s_t, a_t+n_t)$} \Comment{rewards}
        \State {$s_{t+1} \gets f_{\rho}(s_t, a_t+n_t)$} \Comment{system dynamics}
    \EndFor
    \State{\textbf{end for}}
    \State{$\mathcal{R} \gets \mathcal{R} + \gamma^H V_{\bar \psi}(s_H)$}
    \Comment{add terminal state reward}
    \State{$\mathcal{C}= -\mathcal{R}$} 
    \Comment{convert to cost minimization}
    \State {$\beta \gets  min[\mathcal{C}]$}
    \State{$\eta \gets \sum_{i=1}^{N} \exp(-\frac{1}{\lambda}[\mathcal{C}_i- \beta])$}
    
    \State{$w(n) \gets \frac{1}{\eta} \exp(-\frac{1}{\lambda}[\mathcal{C}-\beta])$}
    \Comment {import. sampling weight}
    
    \State{$a_{MPR} \gets a + \sum_{i=1}^{N} w(n_i) n_i$}
    \Comment{adjust action sequence}
    
    \State {$s_{MPR} \gets f_{\rho}(s_0, a_{MPR})$} \Comment{state following optimal action}
        
    
\end{algorithmic}
\end{algorithm}
 \end{minipage}
 \vspace{-0.7cm}
   \end{figure}

\section{Model Predictive Actor-Critic}
We introduce Model Predictive Actor-Critic (MoPAC), an algorithm that leverages the theoretical guarantees provided by Theorem \ref{theorem:mpc} and Lemma \ref{lemma:2}. MoPAC has three main interacting components: (i) model learning from environment transitions, (ii) model predictive rollouts for acquiring samples from optimal trajectories, and (iii) soft updates using a maximum entropy objective for policy learning over a mixture of model and environment data. Algorithm~\ref{Alg:main} summarizes our approach.



\noindent\textbf{Model learning.} We use an ensemble of $N$ functions to approximate the model of the environment $\{f_{\rho_1}, ..., f_{\rho_N}\}$. Each of these functions is a probabilistic deep neural network whose purpose is to approximate the system dynamics, namely the next state of the agent given the current state and action. Probabilistic model ensembles have been studied as a way of realizing Bayesian neural networks, for capturing the epistemic uncertainty when learning complex dynamics, in order to mitigate overfitting when using a single model~\cite{chua2018deep}. Specifically, an ensemble of models is randomly initialized within an observable space; hence, each model learns a different mapping of the dynamics. For generating predictions from the ensembles, we sample transitions from the elite networks, i.e. those with the lowest L2-loss in a validation set, during the model rollouts by sampling uniformly a model for every simulation.

\noindent\textbf{Model predictive rollouts.}
Solving the MPC optimization problem of (\ref{eq:mpc}) is prohibitively expensive and hard to obtain online. This limitation is overcome by the information theoretic model predictive path integral (i-MPPI) method~\cite{williams2017information}, a sampling-based algorithm that uses an approximation of the true dynamics, and is able to optimize both convex and non-convex cost criteria, thus being applicable to large classes of stochastic systems and representations. i-MPPI uses the free energy of the system and relative entropy (KL-divergence) for providing generalized path integral expressions. Crucially, it uses importance sampling for acquiring the optimal control paths. In MoPAC, we use a similar setting for generating model-based rollouts that will profit from the optimal performance guarantees of Sec. \ref{sec:bounds}. In our model predictive rollouts (MPR), we initialize the control sequence employing the learned policy $\pi_{\theta}$ to sample initial actions. Then, we design a similar setup as in i-MPPI for creating trajectory simulations in a specific time horizon $H$, starting from an initial true state (Alg. \ref{alg:mppi}). We evaluate the rewards of the rollouts, using also a VF for the final state \eqref{eq:mpc_bound_2}, and we perform importance sampling of the optimal transitions over all simulations, along with an information-theoretic update of the actions' exploration noise. Finally, we collect the resulting optimal transitions in the model-based replay memory $D_{model}$.

When the approximation error of the learned model is low, the trajectory optimization on the learned model performed by MoPAC yields comparable performance as acting on the real environment (Lemma \ref{lemma:2}). This remark, combined with the use of the VF, allows us to obtain near-optimal performance--up to an approximation error (Theorem \ref{theorem:mpc}). Moreover, using the learned policy distribution for sampling the initial control sequence, together with the relative entropy objective, guarantees that we are simulating transitions in a reasonable area around the learned policy, providing good trajectories for exploitation, while the underlying actor-critic explores new transitions.

\noindent\textbf{Soft policy optimization.} As the underlying actor-critic algorithm in MoPAC, we adopt SAC~\cite{haarnoja2018soft} to benefit from the exploration induced by the soft policy updates based on the maximum entropy principle, counterbalancing the effect of the exploitation induced by MPRs. Nevertheless, MoPAC can be applied to any known off-policy actor-critic algorithm. In SAC, the training of the policy alternates between a soft policy evaluation step based on the soft Bellman backup operator \cite{haarnoja2018softv2}, and a soft policy improvement step that minimizes the expected KL divergence: $J_{\pi}(\theta, D)=\mathbb{E}_{s_t \sim D} [ D_{KL}(\pi|| \exp{\{Q^{\pi}-V^{\pi}\}} )]$, where $Q^{{\pi}}\mbox{, } V^{\pi}$ are the soft Q-function and soft VF of policy $\pi$ respectively.

\section{Experimental results}
 \begin{figure*}
 \centering
 \includegraphics[width = 0.95\textwidth]{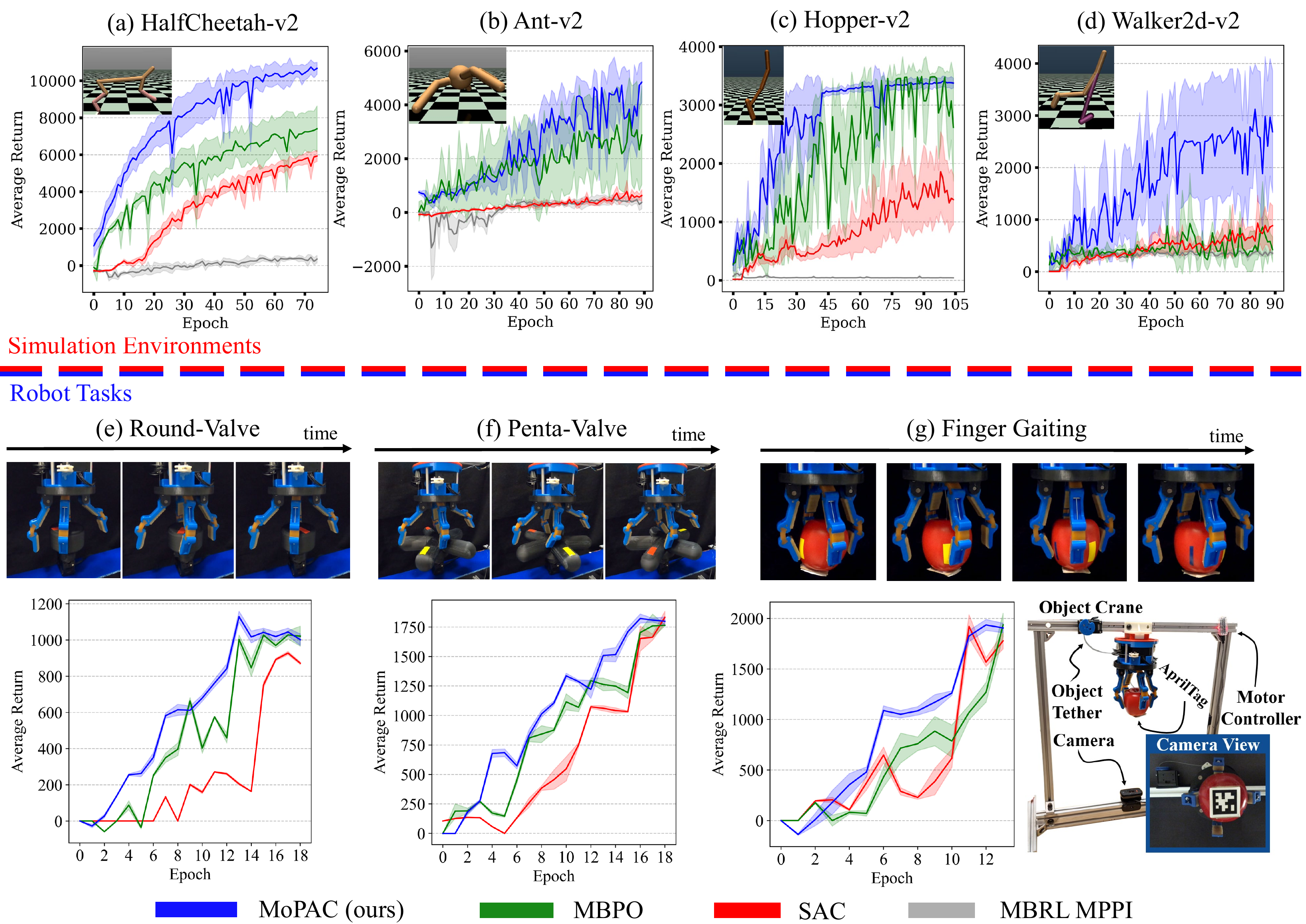}
 \caption{Experimental results comparing MoPAC to other baselines in both MuJoCo environments (a-d) and robotic manipulation tasks (e-g). Results on MuJoCo tasks are averaged over $5$ separate experiments, while for the robotic manipulation tasks we learn a single policy and average its performance on $5$ evaluation runs. All plots show the $95\%$ confidence intervals. The automated object reset system is depicted in the bottom right.}
 \label{Results} 
 \vspace{-0.4cm}
 \end{figure*}
 
\subsection{Simulated tasks}
We evaluate MoPAC in the simulated control tasks of MuJoCo \cite{todorov2012mujoco} included in the OpenAI-gym library~\cite{gym}: \texttt{HalfCheetah-v2}, \texttt{Ant-v2}, \texttt{Hopper-v2}, and \texttt{Walker2d-v2}. We compare the average return of MoPAC over $5$ trials, consisting of $1,000$ true environment interactions per epoch, against the baselines SAC \cite{haarnoja2018softv2}, MBPO \cite{janner2019trust}, and MBRL \cite{williams2017information}. We decided not to compare with \cite{clavera2020model} as the absence of code and a detailed algorithm, makes the reproduction of their results prohibitive.
Here, SAC and MBPO are trained according to the hyperparameter settings provided by their respective works, while for MBRL we use the same settings as MoPAC, namely using horizons $5-15$ with linear annealing for all tasks. The number of environment interactions per episode is constant across all algorithms. A batch size of $10,000$ is chosen for the model rollouts in both MoPAC and MBPO.

Fig.~\ref{Results} shows the average return of each algorithm per environment, plotted w.r.t. the number of epochs. In all environments MoPAC and MBPO outperform both SAC, evincing the advantage of using model rollouts, and MBRL, that strongly suffers from poor exploration in the use of the learned model. Notably, MoPAC learns \textit{faster} than MBPO. Specifically, we observe a significant speed-up in the learning of \texttt{HalfCheetah-v2} and \texttt{Walker2d-v2}. The \texttt{Ant-v2} and \texttt{Hopper-v2} are more challenging tasks, as they require more interactions with the environment to learn their dynamics; in \texttt{Hopper-v2} we observe a speedup in learning and convergence, while in \texttt{Ant-v2} MoPAC learns faster in the first epochs and ends up with slightly better performance than MBPO.
We expect that by further tuning the prediction horizons of our MPR, as now we use the same horizons across all tasks regardless of task-complexity, will result in increased performance.

\subsection{Robotic tasks}
We further underscore the efficacy of MoPAC by comparing our algorithm with SAC and MBPO on a Yale Openhand Model Q \cite{ma2014gaiting, openhand} through two different manipulation tasks--valve rotation and finger gaiting. The Model Q is an open source underactuated hand with four two-link fingers and four total actuators (Dynamixel XM-430). Within the hand, a single motor actuates two opposing fingers that are coupled by a differential, allowing passive reconfigurability between the fingers. Two additional motors actuate the remaining two fingers individually. The fourth and final motor serves to rotate the palm of the hand, allowing the two coupled fingers to reorient perpendicular to the palm axis (Fig. \ref{Splash}). Being underactuated, the Model Q's joint configuration cannot be accurately determined, as with many soft, compliant, or underactuated hands, this hand is not equipped with joint encoders or tactile sensors \cite{morgan2021towards}. This inherent compliance suits the spirit of our evaluations well, as the reconfigurability of the fingers presents added challenges in sufficiently learning the dynamics of the environmental interactions.

\noindent\textbf{Valve Rotation.}
We develop two different valves, the penta-valve and the round-valve (resembling the geometry of a door knob), for evaluation (Fig. \ref{Splash}). Each of the valves were connected to a current-disabled Dynamixel actuator placed directly below the hand as to measure rotation via the encoder. Albeit disabled, valve rotation was resisted by the $353.5:1$ gearbox inside of the actuator, generally limiting rotation to only large actions from the hand (see twisting of flexure joints in Fig. \ref{Splash}). Policies were trained for both valves with each of the three aforementioned algorithms on the physical system. The goal of this task was to continually rotate the valve counterclockwise until episode completion ($50$ actions). We define a standardized reward function, $r(\cdot) = \theta_{s'} - \theta_{s}$, and run each of the algorithms with similar hyperparameters and epoch lengths ($50$ interactions per episode, $5$ episodes per epoch, MPR horizons $2-5$ annealed).

Experimental results to these tasks, depicted in Fig. \ref{Results}(e-g), illustrate the benefit of our MoPAC algorithm compared to baselines. Notably, the reward convergence of the round-valve was less than that of the penta-valve, generally due to the increased difficulty in estimating the task's dynamical hand-object nature, since the round-valve requires grasping before rotation. Due to the design of the hand's flexure joints, the motors are decreasingly able to provide rotational torque the more a finger is actuated, i.e. off-axis torsion limits the amount of force the finger-contacts can transmit to rotate the object. This further leads to a grasping and rotational action that, if enacted with rigid joints instead of compliant ones, would result in a reward from the valve rotation, but does not necessarily happen with flexure joints. Generally, the function of mapping grasping force and palm orientation to valve rotation must first be inferred by the dynamics model in order to sufficiently learn the task. 

Albeit in some ways easier to estimate, hand-object dynamics of the penta-valve is also not deterministic, as the fingers interact with both the fingerpad (rubber) and side of the finger (plastic), where slipping can occur. This task noted a higher converging reward comparatively, as it was not required that the valve be grasped for rotation, dismissing off-axis torsion constraints. Towards the beginning of training, the hand utilized the two independent, non-rotational fingers to achieve a greedy reward, i.e. small movement given the valve configuration, but soon thereafter noted that retreating the fingers into a non-actuated position generally benefited the cumulative reward as to limit interference with the valve. 

\noindent\textbf{Finger Gaiting.}
In addition to the valve rotation assessment, we further challenged our algorithm with the task of finger gaiting--a task requiring coordinated movement between the opposing finger pairs in the hand. Here, the hand is not only expected to maintain a stable grasp on the object during the entire episode, but to rotate the object about the palm axis of the hand (Fig. \ref{Splash}). We incorporated a similar reward function as in the previous section, $r(\cdot) = \theta_{s'_{palm}} - \theta_{s_{palm}}$, but now the object is not constrained to only reorient along the reward axis, being free to move in $SE(3)$ space. In this evaluation, we utilize the apple (Obj. \#13) from the YCB Object and Model Set \cite{calli2015benchmarking} as to maintain standardization of the task. 

The episode starts with a stable grasp acquired by the two individually driven fingers, with the coupled fingers close to, but not touching, the object. We monitor the pose of the object during manipulation via an AprilTag \cite{wang2016apriltag} attached to the bottom of the apple, and detected by an external camera. Subsequently, the reward of an action can be calculated through this setup, in addition to detecting when the hand drops the object. When a drop is detected, the episode is ended and the object is systematically reset via an automated object reset system (Fig. \ref{Results}). This system consists of an object crane that controls a tether routed through the center of the hand. During reset, the object lifts the object into the palm of the hand as to stabilize, and then slowly lowers the object into the starting position. 

In this evaluation, we note similar curves to those in the valve rotation task (Fig. \ref{Results}). Specifically, we see MoPAC outperform the two other algorithms, reaching convergence faster with fewer environment interactions. Intuitively, this quality is increasingly advantageous for tasks where system reset cannot be completed quickly or autonomously--although it was possible in this experiment, object reset required about $12$ seconds and slowed the learning process. Training for the valve rotation tasks took $\sim 2$ hours for each valve and each learning algorithm, whereas training for finger gaiting took $\sim 5$ hours. Notably, if we were to stop training upon MoPAC convergence, significant time can be saved, which is especially desirable when increasing task complexities. 

\section{Conclusions}
Transferring the advances of deep RL into real-world robotic problems is challenging. Deep model-free RL~(MFRL) methods, though able to learn complex skills, typically require an excessive amount of interactions with the environment, making their applicability prohibitive for robotics. On the other hand, model-based RL~(MBRL) approaches learn the dynamics model and select actions through trajectory optimization techniques; albeit sample-efficient, they suffer from local optima. In this work, we proposed Model Predictive Actor-Critic (MoPAC), a method seeking to combine the advantages of deep MF actor-critic methods with the data-efficiency of MB approaches. In particular, MoPAC introduces model predictive rollouts, inspired by information-theoretic model predictive path integrals, based on the principles of the dynamics free-energy and on an information-theoretic constraint for collecting samples through trajectory optimization. MoPAC uses a MFRL actor-critic algorithm for policy improvement and model learning, and it uses the model for performing the model predictive rollouts to collect additional samples for guiding policy learning. The core advantage of MoPAC is that, though MB rollouts favor policy exploitation through planning using the model, the MFRL actor-critic encourages efficient exploration for policy optimization and model learning.

Our model predictive rollouts are backed up by a performance bound, which guarantees that sufficiently low errors in the value function, and model approximation, yields near-optimal performance. We empirically demonstrate the efficiency of MoPAC in providing accelerated policy learning for simulated control tasks against representative baselines. Furthermore, we showcase the applicability of MoPAC for learning challenging in-hand manipulation tasks with a four-fingered robotic hand. In the future, we will study ways of principally adjusting the mixing of the MB-MF samples across the training process, but also ways of scheduling the MB over the MF method according to the learning progress, and vice versa.






\section{Appendix}
\textit{Proof of Theorem 1.}
Let the performance of applying policy $\hat \pi$ from MPC using the approximated model $\hat f$ be denoted as $\hat V$ and the performance gain of applying the optimal policy $\pi^{*}$ on the perfect model $f^*$ be $V^*$ over a planning horizon $H$. The performance error for any given starting state $s$ is
\small
\begin{align} \label{eq:proof1}
    V^*(s) -\hat V(s) &=
    \sum_{s \sim f^*}[\sum_{t=0}^{H-1}f^*(s_t,a_t)\gamma^t r_t + \gamma^H V^*(s_H)] \nonumber \\
    &-\sum_{s \sim \hat f}[\sum_{t=0}^{H-1}\hat f(s_t,a_t)\gamma^t r_t + \gamma^H \hat V(s_H)].
\end{align}
\normalsize
\noindent Adding and subtracting \small $\sum_{s \sim \hat f}[\sum_{t=0}^{H-1}\hat f(s_t,a_t)\gamma^t r_t + \gamma^H V^*(s_H)]$\normalsize in \eqref{eq:proof1} yields
\small
\begin{align} 
     V^*(s) -\hat V(s) &=  \gamma^H\sum_{s \sim \hat f} [V^*(s_H) - \hat V(s_H)] \nonumber  \\
    &\quad +\sum_{s \sim f^*}[\sum_{t=0}^{H-1}f^*(s_t,a_t)\gamma^t r_t + \gamma^H V^*(s_H)] \nonumber \\
    &-\sum_{s \sim \hat f}[\sum_{t=0}^{H-1}\hat f(s_t,a_t)\gamma^t r_t + \gamma^H V^*(s_H)]. \label{eq:proof2}
\end{align}
\normalsize

\noindent Since the value function error is upper-bounded by $max_s|V^*(s)-\hat V(s)|=\epsilon_V$, we can bound the following relations as
\small
\begin{align}\label{eq:proof3}
    \sum_{s \sim f^*}&[\sum_{t=0}^{H-1}f^*(s_t,a_t)\gamma^t r_t + \gamma^H V^*(s_H)] \leq \nonumber \\
    \sum_{s \sim f^*}&[\sum_{t=0}^{H-1}f^*(s_t,a_t)\gamma^t r_t + \gamma^H \hat V(s_H)] + \gamma^H \epsilon_V& \\
    \sum_{s \sim \hat f}&[\sum_{t=0}^{H-1}\hat f(s_t,a_t)\gamma^t r_t + \gamma^H V^*(s_H)] \geq \nonumber \\
    \sum_{s \sim \hat f}&[\sum_{t=0}^{H-1}\hat f(s_t,a_t)\gamma^t r_t + \gamma^H \hat V^(s_H) - \gamma^H \epsilon_V.
\end{align}
\normalsize
Substituting \eqref{eq:proof3} into \eqref{eq:proof2}, we have 
\small
\begin{align}
      &V^*(s) - \hat{V}(s) \leq \gamma^H\sum_{s \sim \hat f} [V^*(s_H) - \hat V(s_H)] + 2\gamma^H \epsilon_{V} \nonumber \\
      &+ \sum_{s \sim f^*}[\sum_{t=0}^{H-1}\hat f(s_t,a_t)\gamma^t r_t -
      \sum_{s \sim \hat f}[\sum_{t=0}^{H-1}\hat f(s_t,a_t)\gamma^t r_t
      \nonumber \\ &\leq \gamma^H\sum_{s \sim \hat f} [V^*(s_H) - \hat V(s_H)] + 2\gamma^H \epsilon_V \nonumber \\
     &+ r_{max} \{\sum_{s \sim f^*}[\sum_{t=0}^{H-1}\hat f(s_t,a_t)\gamma^t] -  \sum_{s \sim \hat f}[\sum_{t=0}^{H-1}\hat f(s_t,a_t)\gamma^t] \} \nonumber \\ &= \gamma^H\sum_{s \sim \hat f} [V^*(s_H) - \hat V(s_H)] + 2\gamma^H \epsilon_V + r_{max} \sum_{t=0}^{H-1} \gamma^t \epsilon_f \nonumber \\ 
     &\leq 2 \gamma^H \epsilon_V(1 + \gamma^H + \gamma^{2H} + ...) + r_{max}\sum_{t=0}^{H-1} \gamma^t \epsilon_f \nonumber \\
     &\leq \frac{2 \gamma^H\epsilon_V}{1-\gamma^H} + r_{max}\sum_{t=0}^{H-1} \gamma^t \epsilon_f = \frac{2 \gamma^H\epsilon_V}{1-\gamma^H} + r_{max}\frac{1-\gamma^H}{1-\gamma} \epsilon_f \label{eq:proof4}\\  
     &\leq  \frac{r_{max}\epsilon_f}{1-\gamma}.\label{eq:proof5}
\end{align}
\normalsize
For MPC with imperfect dynamics and prediction horizon $H$, the bound \eqref{eq:proof4} holds, showcasing the inevitable error due to the model approximation. However, this relation can be upper-bounded by the quantity in \eqref{eq:proof5} considering an infinite-horizon, $H\to \infty$, prediction. The bound of \eqref{eq:proof4} shows that sufficiently low approximation errors in the VF and the model can yield near-optimal performance, which is related to the prediction horizon and the discounted factor.  


\clearpage
\bibliographystyle{IEEEtran}
\bibliography{MPAC}

\end{document}